\documentclass[11pt,a4paper]{article}
\usepackage[hyperref]{emnlp2020}
\usepackage{times}
\usepackage{latexsym}
\usepackage{booktabs}       
\usepackage{pgfplots}
\usepackage{subcaption}
\usepackage{amsmath}
\usepackage{enumitem}

\newcommand{\scifact}{\textsc{SciFact}\xspace}
\newcommand{\vs}{\textsc{VeriSci}\xspace}
\newcommand{\vt}{\textsc{VerT5erini}\xspace}
\newcommand{\supports}{\textsc{Supports}\xspace}
\newcommand{\refutes}{\textsc{Refutes}\xspace}
\newcommand{\noinfo}{\textsc{NoInfo}\xspace}
\usepackage{microtype}
\usepackage{xspace}
\aclfinalcopy 

\newcommand{\red}[1]{\textcolor{red}{#1}}
\newcommand{\green}[1]{\textcolor{green}{#1}}
\newcommand{\yellow}[1]{\textcolor{yellow}{#1}}

\title{Scientific Claim Verification with \vt}

\author{%
  Ronak Pradeep, Xueguang Ma, Rodrigo Nogueira, and Jimmy Lin \\[1ex]
  David R. Cheriton School of Computer Science\\
  University of Waterloo\\
}

\date{}

\begin{document}
\maketitle
\begin{abstract}
This work describes the adaptation of a pretrained sequence-to-sequence model to the task of scientific claim verification in the biomedical domain. 
We propose \vt that exploits T5 for abstract retrieval, sentence selection and label prediction, which are three critical sub-tasks of claim verification. 
We evaluate our pipeline on \scifact, a newly curated dataset that requires models to not just predict the veracity of claims but also provide relevant sentences from a corpus of scientific literature that support this decision.
Empirically, our pipeline outperforms a strong baseline in each of the three steps.
Finally, we show \vt's ability to generalize to two new datasets of COVID-19 claims using evidence from the ever-expanding CORD-19 corpus.
\end{abstract}

\section{Introduction}

The popularity of social media and other means of disseminating content, combined with automated algorithms that amplify signals, has increased the proliferation of misinformation.
This has caused increased attention in the community towards building better fact verification systems.
Until recently, most fact verification datasets were constrained to domains such as Wikipedia, discussion blogs, and social media \cite{thorne-etal-2018-fever, hanselowski-etal-2019-richly}.

In the current environment, amidst the COVID-19 pandemic and the unease that comes perhaps with insufficient insight about the virus, there has been a sharp increase in scientific curiosity among the general public.
While such curiosity is always appreciated, this has inadvertently resulted in a large spike of scientific facts being misrepresented, often to push personal or political agenda, inducing ineffective and often even harmful policies and behaviours.

To mitigate this issue, \citet{Wadden2020FactOF} introduced the task of scientific claim verification where systems need to evaluate the veracity of a claim against a scientific corpus. 
To facilitate this, they introduced the \scifact dataset that consists of scientific claims accompanied with abstracts that either support or refute the claim.
The dataset also provide a set of rationale sentences for each claim that is often both necessary and sufficient to conclude its veracity.

In addition, they provide \vs, a baseline for this task that takes inspiration from previous state of the art systems \cite{deyoung-etal-2020-eraser} for the FEVER claim verification dataset \cite{thorne-etal-2018-fever}.
This pipeline retrieves relevant abstracts by TF-IDF similarity, uses a BERT-based model \cite{devlin-etal-2019-bert} to select rationale sentences, and finally label each abstract as either \supports, \noinfo or \refutes with respect to the claim.

\begin{figure*}[t]
\begin{center}
\centerline{\includegraphics[width=0.98\textwidth]{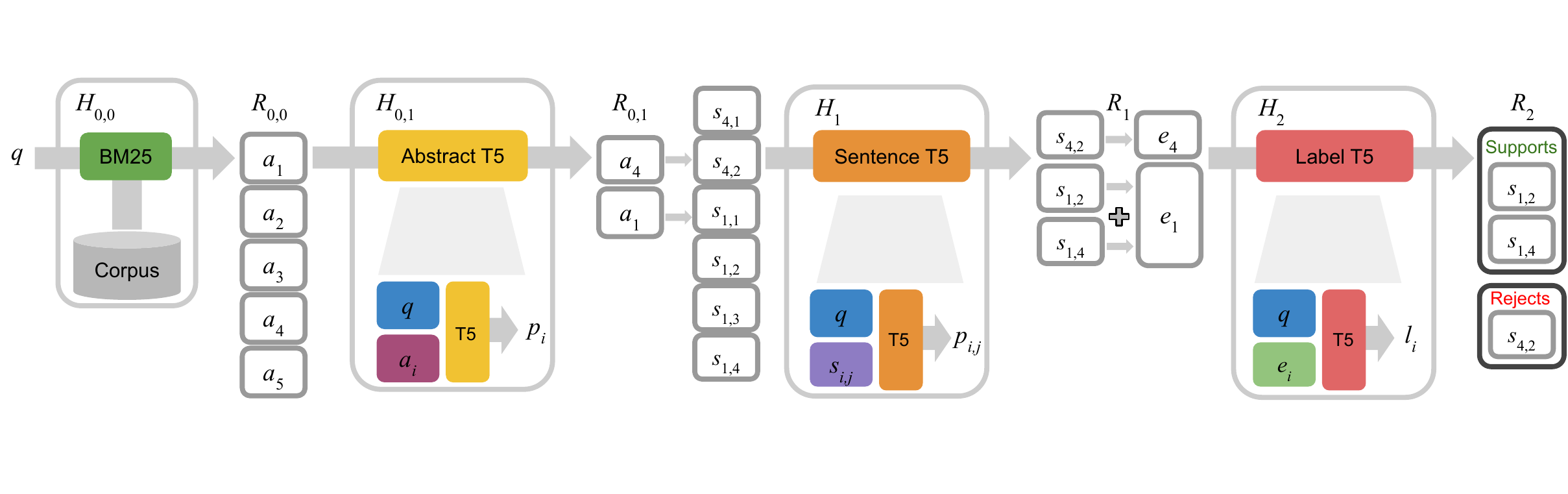}}
\vspace{-0.1cm}
\caption{Illustration of the \vt pipeline.} 
\label{fig:overview}
\end{center}
\end{figure*}

Despite the success of BERT for a tasks like passage-level \cite{Nogueira_etal_arXiv2019}, document-level \cite{dai2019contextaware, macavaney2019cedr, yilmaz2019cross} and sentence-level \cite{soleimani2019bert} retrieval, there is evidence that ranking with sequence-to-sequence models can achieve even better effectiveness, particularly in zero-shot scenarios or with limited training data \cite{Nogueira_etal_FindingsEMNLP2020}.
This was further demonstrated in the TREC-COVID challenge~\cite{roberts2020trec} where one of the best performing systems used sequence-to-sequence models for retrieval~\cite{zhang2020covidex}.
Similar trends are noted in CovidQA \cite{tang2020rapidly}, a question answering dataset for COVID-19, where zero-shot sequence-to-sequence models outperformed other baselines.

Hence, we propose \vt, where all three steps---abstract retrieval, sentence selection and label prediction exploit T5 \cite{raffel2019exploring}, a powerful sequence-to-sequence language model.
\vt significantly outperforms the \vs baseline on the \scifact tasks and hence qualifies as a strong pipeline for the task of Scientific Claim Verification.
We also demonstrate the efficacy of our system in verifying two different sets of COVID-19 claims with no additional training or hyperparameter tuning.

\section{Task}
In the \scifact task, systems are provided with a scientific claim $q$ and a corpus of abstracts $\mathcal{C}$ and tasked to return:

\begin{itemize}[leftmargin=*]
    \item A set of evidence abstracts $\hat{\mathcal{E}}(q)$.
    \item A label $\hat{y}(q, a)$ that maps claim $q$ and abstract $a$ to one of $\{\supports, \refutes, \noinfo\}$.
    \item A set of rationale sentences $\hat{S}(q, a)$ when $\hat{y}(q, a) \in \{\supports, \refutes\}$.
\end{itemize}

Given the ground truth label $y(q,a)$, the set of gold abstracts $\mathcal{E}(q)$, and the set of gold rationales $\mathcal{R}(q,a)$ (each gold rationale is a set of sentences), the predictions are evaluated in two ways:
\begin{itemize}[leftmargin=*]
    \item  \textbf{Abstract-level evaluation}, where systems are judged on whether they can identify abstracts that support or refute the claim.
    First, $a \in \hat{\mathcal{E}}(q)$ is \emph{correctly labelled} if both $a \in \mathcal{E}(q)$ and $\hat{y}(q, a) = y(q,a)$.
    Second, it is \emph{correctly rationalized}, if in addition, $\exists R \in \mathcal{R}(q,a)$ such that $R \subseteq \hat{S}(q, a)$.\footnote{In \scifact's abstract-level evaluation, it is required that $|\hat{S}(q, a)| \leq 3$.} These evaluations are referred to as Abstract\textsubscript{Label-Only} and Abstract\textsubscript{Label+Rationale}, respectively.
    \item \textbf{Sentence-level evaluation}, where systems are evaluated on whether they can identify sentences sufficient to justify the abstract-level predictions. 
    First, $\hat{s} \in \hat{S}(q,a)$ is \emph{correctly selected} if $\exists R \in \mathcal{R}(q,a)$ such that both $\hat{s} \in R$ and $R \subseteq \hat{S}(q,a)$. 
    Second, it is \emph{correctly labelled}, if in addition, $\hat{y}(q, a) = y(q,a)$. These evaluations are referred to as Sentence\textsubscript{Selection-Only} and Sentence\textsubscript{Selection+Label}, respectively.
\end{itemize}

\begin{table*}[t]
\centering\centering\resizebox{1\textwidth}{!}{
\begin{tabular}{p{3cm}|p{2cm}|p{12cm}}
\toprule
\textbf{Claim} & \textbf{Label} & \textbf{Evidence} \\
\toprule
ALDH1 expression is associated with poorer prognosis in breast cancer.
& \green{\supports} &
Application of stem cell biology to breast cancer research has been limited by the lack of simple methods for identification and isolation of normal and malignant stem cells.
$\ldots$
\textbf{In a series of 577 breast carcinomas, expression of ALDH1 detected by immunostaining correlated with poor prognosis.}
$\ldots$
\\
\midrule
CX3CR1 on the Th2 cells impairs T cell survival
& \red{\refutes} & 
Allergic asthma is a T helper type 2 (T(H)2)-dominated disease of the lung.
$\ldots$
\textbf{We found that CX3CR1 signaling promoted T(H)2 survival in the inflamed lungs, and injection of B cell leukemia/lymphoma-2 protein (BCl-2)-transduced CX3CR1-deficient T(H)2 cells into CX3CR1-deficient mice restored asthma.}
$\ldots$\\
\midrule
Arterioles have a larger lumen diameter than venules.
& \yellow{\noinfo} & 
N/A\\
\bottomrule
\end{tabular}
}
\vspace{-0.1cm}
\caption{Three \scifact claims, their labels and their corresponding evidence (rationale highlighted in \textbf{bold}) if available. \vt correctly predicts these labels and retrieves the matching evidence.}
\vspace{-0.25cm}
\label{tab:scifact_examples}
\end{table*}

\section{Method}
Our proposed framework, \vt (see Figure~\ref{fig:overview}), has three major components:
\begin{enumerate}[leftmargin=*]
    \item \textbf{$H_0$: Abstract Retrieval} --- which given claim $q$ retrieves the top-$k$ abstracts from corpus $\mathcal{C}$.
    \item \textbf{$H_1$: Sentence Selection} --- which given claim $q$ and one of the top-$k$ abstracts $a$, selects sentences from $a$ that form  $\hat{S}(q, a)$.
    \item \textbf{$H_2$: Label Prediction} --- which given claim $q$ and the rationale sentences $\hat{S}(q, a)$, predicts the final label $\hat{y}(q,a)$.
\end{enumerate}

\subsection{$H_0$: Abstract Retrieval}
Given a scientific claim $q$ and a corpus $\mathcal{C}$ of scientific abstracts, $H_0$ is tasked with retrieving the top-$k$ abstracts from $\mathcal{C}$.
We propose both a single-stage and a two-stage abstract retrieval pipeline.

In both cases, the first stage $H_{0,0}$ involves treating the query as a "bag of words" for ranking abstracts from the corpus using a BM25 scoring function \citep{robertson1995okapi}.
Our implementation uses the Anserini IR toolkit~\cite{Yang_etal_SIGIR2017,Yang_etal_JDIQ2018},\footnote{\url{http://anserini.io/}} which is built on the popular open-source Lucene search engine.
The output of this stage is a list of $k_{0}$ candidate abstracts.

The second abstract reranking stage, $H_{0,1}$, is tasked to estimate a score $p$ quantifying how relevant a candidate abstract $a$ is to a query $q$.
In this stage, the abstracts retrieved in $H_{0,0}$ are reranked by a pointwise reranker, which we call monoT5.
Our reranker is based on \citet{Nogueira_etal_FindingsEMNLP2020}, which uses T5~\cite{raffel2019exploring}, a sequence-to-sequence model pretrained with a similar masked language modeling objective as BERT.
In this model, all target tasks are cast as sequence-to-sequence tasks.
We adapt the approach to abstract reranking by using the following input sequence:
\begin{align*}
\text{Query: } q \ \ \text{ Document: } a \ \ \text{ Relevant:}
\end{align*}
The model is fine-tuned to produce the words ``true'' or ``false'' depending on whether the abstract is relevant or not to the query.
That is, ``true'' and ``false'' are the ``target words'' (i.e., ground truth predictions in the sequence-to-sequence transformation).
Since \scifact abstracts tend to be longer than context limit of T5, we first segment each abstract into spans by applying a sliding window of 6 sentences with a stride of 3. 

In order to fine-tune monoT5 on abstract reranking in \scifact, we use all cited abstracts in the train set as positive examples.
For each claim, we select negative examples by randomly selecting a non-ground truth abstract among the top-10 BM25 ranked candidates.
We train on this set with a batch size of 128 for 200 steps which corresponds to approximately 5 epochs.

At inference time, we first to compute probabilities for each query--segment pair (in a reranking setting), we apply a softmax only on the logits of the ``true'' and ``false'' tokens.
We then obtain the relevance score of the document as the highest probability assigned to the ``true'' token among all segments.
The top-$k_0$ abstracts, $R_0$, with respect to these scores are then selected.

We run inference with three different monoT5\footnote{All models are T5-3B.} settings for abstract reranking: (1) fine-tuned on the \href{https://microsoft.github.io/MSMARCO-Passage-Ranking/}{MS MARCO passage}~\cite{nguyen2016ms} dataset; (2) fine-tuned on MS MARCO then fine-tuned again on the medical subset of MS MARCO~\cite{macavaney:arxiv2020-sledge}; and (3) fine-tuned on MS MARCO then fine-tuned again on \scifact.

We choose to pretrain relevance classifiers on MS MARCO passage as it has been shown to help in various other tasks~\cite{Nogueira_etal_FindingsEMNLP2020, zhang2020covidex, yilmaz2019cross}. 
Similarly, \citet{macavaney:arxiv2020-sledge} demonstrate that fine-tuning the classifiers on MS MARCO MED helps with biomedical-domain relevance ranking.
\subsection{$H_1$: Sentence Selection}
In this stage, the goal is to select rationale sentences $\hat{S}(q,a)$ from each abstract $a$ for each of the top-$k$ abstracts retrieved $\hat{\mathcal{E}}(q)$.
We use T5 for this task too.
The following input sequence is used:
\begin{align*}
\text{Query: } q \ \ \text{ Document: } s \ \ \text{ Relevant:}
\end{align*}
\noindent where $s$ is a sentence in the abstract $a$. 

We fine-tune a monoT5 (trained on MS MARCO passage) on \scifact's gold rationales as positive examples and sentences randomly sampled from $\mathcal{E}(q)$ as negatives.
We train on this set of sentences with a batch size of 128 for 2500 steps.

During inference, similar to abstract ranking, we compute a probability of the sentence being relevant based on the logits of ``true'' and ``false'' tokens.
Finally, we filter out all sentences whose ``true'' probability is below the threshold of $0.999$ to obtain $\hat{S}(q,a)$.

\subsection{$H_2$: Label Prediction}
Given the claim $q$, an abstract $a$ and their corresponding set of rationale sentences $\hat{S}(q,a)$, $H_2$ is tasked to predict a label $\hat{y}(q, a) \in \{\supports, \noinfo, \refutes\}$.
Yet again, we use T5 for this task with input sequence:
\begin{align*}
    \text{hypothesis: } q \ \text{ sentence$1$: } s_1 \ \cdots \ \text{ sentence$z$: } s_{z} 
\end{align*}
\noindent where $s_1,\cdots,s_z$ are the rationale sentences in $\hat{S}(q,a)$.
The target sequence is one of ``true'', ``weak'' or ``false'' tokens corresponding to the labels \supports, \noinfo or \refutes, respectively.

\supports and \refutes training examples are selected from evidence sets of cited abstracts for each claim. 
The sentences in each evidence set will be concatenated with the claim in the above input sequence format as a single example for the corresponding label.
The \noinfo examples are selected by concatenating 1 or 2 randomly-selected non-rationale sentences from each of the cited abstracts across all labels.
Here, we fine-tuned a fresh T5-3B (that was just pretrained on the mixture task) and not a fine-tuned monoT5 since there is no natural transfer from the relevance ranking task.
We use a batch size of 128 and pick out the best checkpoint after $[200, 400, 600, 800, 1000]$ steps based on the development set scores.

During inference, the token with the highest probability will be the label $\hat{y}(q,a)$ for abstract $a \in \mathcal{E}(q)$.

\section{Experimental Setup}
\subsection{Datasets}

\begin{table}[t]
\centering\centering\resizebox{0.49\textwidth}{!}{
\small
\begin{tabular}{l|cccc}
\toprule
Set   & \supports & \noinfo & \refutes & Total\\
\toprule
Train  & 332 & 304 & 173 & 809\\
Dev & 124 & 112 & 64 & 300\\
Test & 100 & 100 & 100 & 300\\

\bottomrule
\end{tabular}
}
\vspace{-0.1cm}
\caption{\scifact label distribution.}
\label{tab:scifact_distribution}

\vspace{-0.25cm}
\end{table}
\smallskip \noindent {\bf \scifact}~\cite{Wadden2020FactOF} consists of a corpus of 5,183 abstracts.
Abstracts that support or refute each claim are annotated with rationale sentences (see Table~\ref{tab:scifact_examples} for examples).
The label distribution is provided in Table~\ref{tab:scifact_distribution}.
There are 1,409 claims, 809 of which are part of the training set and the rest are split equally across the development and test set. 
The test set is balanced with 100 claims for each class (\supports, \noinfo and \refutes).
Yet from Table~\ref{tab:scifact_distribution}, it is clear that the training and development sets have significant class imbalance.
This coupled with the small set sizes highlight the importance of zero-shot or few-shot systems for this task.

To show that our system is
able to verify claims related to COVID-19 by
identifying evidence from the much larger CORD-19 corpus,\footnote{We use the 2020-06-17 dump of CORD-19, which contains 192,459 abstracts, about 40 times as many abstracts as those in \scifact.} we evaluate \vt in a zero-shot setting on two other datasets:

\begin{table}[t]
\centering\centering\resizebox{0.49\textwidth}{!}{
\small
\begin{tabular}{l|ccc}
\toprule
Claim Set   & \supports & \refutes & Total\\
\toprule
COVID-19 \scifact  & - & - & 36\\
\midrule
COVID-19 Scientific & 41 & 101 & 142\\
\bottomrule
\end{tabular}
}
\vspace{-0.1cm}
\caption{COVID-19 claims.}
\label{tab:covid_claims}

\vspace{-0.25cm}
\end{table}
\begin{table*}[t]
\centering\centering\resizebox{1\textwidth}{!}{
\begin{tabular}{p{3.6cm}|p{2cm}|p{12cm}}
\toprule
\textbf{Claim} & \textbf{Label} & \textbf{Evidence} \\
\toprule
Hypertension and Diabetes are the most common comorbidities for COVID-19.
& \green{\supports} &
\textbf{Investigations reported that hypertension, diabetes, and cardiovascular diseases were the most prevalent comorbidities among the patients with coronavirus disease 2019 (COVID-19).
}
$\ldots$
The aim of this review was to summarize the current knowledge about the relationship between hypertension and COVID‐19 and the role of hypertension on outcome in these patients.
\\
\midrule
The Secondary Attack rate of COVID-19 is 10.5\% for household members/close contacts.
& \red{\refutes} & 
Background: As of April 2, 2020, the global reported number of COVID-19 cases has crossed over 1 million with more than 55,000 deaths.
$\ldots$
\textbf{We estimated the household SAR to be 13.8\% (95\% CI: 11.1–17.0\%) if household contacts are defined as all close relatives and 19.3\% (95\% CI: 15.5–23.9\%) if household contacts only include those at the same residential address as the cases, assuming a mean incubation period of 4 days and a maximum infectious period of 13 days.}\\
\bottomrule
\end{tabular}
}
\vspace{-0.1cm}
\caption{Two COVID-19 claims from \scifact, their \textit{predicted} labels and their corresponding \textit{predicted} evidence (rationale highlighted in \textbf{bold}).}
\label{tab:qual_covid19_scifact}
\end{table*}

\begin{table*}[t]
\centering\centering\resizebox{1\textwidth}{!}{
\begin{tabular}{p{3.6cm}|p{2cm}|p{12cm}}
\toprule
\textbf{Claim} & \textbf{Label} & \textbf{Evidence} \\
\toprule
Some people become infected by COVID-19 but don't develop any symptoms and don't feel unwell.
& \green{\supports} &
COVID-19 is an emerging infectious disease with widespread transmission of the coronavirus SARS-CoV-2 in the Netherlands.
$\ldots$
\textbf{Others do not show any symptoms, but can still contribute to the transmission of the virus.}
$\ldots$\\
\midrule
Young people will not get COVID-19.
& \red{\refutes} & 
Objective: To explore the epidemiological characteristics of COVID-19 associated with SARS-Cov-2 in Guizhou province, and to compare the differences in epidemiology with other provinces.
$\ldots$
\textbf{Most of COVID-19 patients were 18-45 years old (52.27\%).}
$\ldots$
\textbf{
CONCLUSION: Among the cases, most patients were young adults.}\\
\midrule
Bill Gates caused the infection of COVID-19.
& \red{\refutes} & 
N/A\\
\bottomrule
\end{tabular}
}
\vspace{-0.1cm}
\caption{Three COVID-19 Scientific claims from \citet{lee2020misinformation}, their \textit{predicted} labels and their corresponding \textit{predicted} evidence (rationale highlighted in \textbf{bold}). Note that if the system can not find any supporting evidence for a claim, it refutes the claim.}
\vspace{-0.25cm}
\label{tab:qual_covid19_scientific}
\end{table*}

\smallskip \noindent {\bf COVID-19 \scifact}~\cite{Wadden2020FactOF} is a set of 36 COVID-related claims curated by a medical student.
In this set, the same claim can sometimes be both supported and refuted by different abstracts, a scenario not observed in the main \scifact task.
Two examples in this set are shown in Table~\ref{tab:qual_covid19_scifact}.

\smallskip \noindent {\bf COVID-19 Scientific}~\cite{lee2020misinformation} contains 142 claims (label distribution in Table~\ref{tab:covid_claims}) gathered by collecting COVID-related scientific truths and myths from sources like  the Centers for Disease Control and Prevention (CDC), MedicalNewsToday, and the World Health Organization (WHO).
Unlike the other two datasets, COVID-19 Scientific only provides a single label $y(q) \in \{\supports, \refutes\}$ for a claim.
They mention that in the construction of the dataset, claims that were unverifiable according to the CDC or the WHO were mapped to \refutes.
Hence, we make the following modifications to \vt:
\begin{itemize}[leftmargin=*]
    \item First if $\hat{y}(q,a) = \noinfo$, then $\hat{y}(q,a)$ is modified to $\refutes$.
    \item Second, $\hat{y}(q) = \max_{a \in \hat{\mathcal{E}}(q)} \hat{y}(q,a)$.
    \item Third, if $|\bigcup_{a \in \hat{\mathcal{E}}(q)} \hat{S}(q,a)| = 0$ i.e. the set of all evidence sentences across the abstracts is empty, then $\hat{y}(q) = \refutes$.
\end{itemize}
Three examples from this set are shown in Table~\ref{tab:qual_covid19_scientific}.
As once can imagine, it would be impossible to find any discussion for outlandish claims like ``Bill Gates caused the infection of COVID-19'' in a corpus of biomedical literature and hence \vt maps it to \refutes.

\subsection{Baselines}
For the \scifact and COVID-19 \scifact end-to-end tasks, the baseline system used is \vs~\cite{Wadden2020FactOF}.
It has an abstract retrieval module that uses TF-IDF, a sentence selection module trained on \scifact and a label prediction module trained on FEVER + \scifact.
For the abstract retrieval module, they note the best full-pipeline development set scores by retrieving the top three documents.

For the COVID-19 Scientific task, we compare with the following two baselines established by \citet{lee2020misinformation}:
\begin{itemize}[leftmargin=*]
    \item LiarMisinfo~\cite{lee2020misinformation} uses a BERT-large~\cite{devlin-etal-2019-bert} label prediction model fine-tuned on LIAR-PolitiFact~\cite{wang2017liar}, a set of 12.8k claims collected from PolitiFact.
    It is worth noting that LIAR-PolitiFact does not contain any claims related to COVID-19.
    \item LM Debunker~\cite{lee2020misinformation} uses GPT-2~\cite{radford2019language} to determine the perplexity of the claim given evidence sentences.
    Claims with a perplexity score higher than a threshold are labeled \refutes while the others are labeled \supports.
\end{itemize}
The sentence selection module in both baselines employ TF-IDF followed by some rule-based evidence filtering to select the top three sentences for each claim.
LiarMisinfo represents a zero-shot model where no fine-tuning is performed on the COVID-19 Scientific set.
LM Debunker, on the other hand, first partitions the set into a validation and a test set.
Then, the validation set is used to tune the perplexity threshold for the model following which evaluation is performed on the test set.
\section{Results}
\subsection{Abstract Retrieval}
\begin{table}[t]
\small
\centering{
\begin{tabular}{l|ll}
\toprule
Method & R@3 & R@5\\
\toprule
Oracle  & 97.61 & 100.00 \\
TF-IDF & 69.38 & 75.60 \\
\midrule
BM25 & 79.90 & 84.69 \\
T5 (MS MARCO) & 86.12 & \textbf{89.95} \\
T5 (MS MARCO MED)  & 85.65 & 89.00 \\
T5 (\scifact) & \textbf{86.60} & 89.40 \\
\bottomrule
\end{tabular}
}
\vspace{-0.1cm}
\caption{Comparison of abstract retrieval methods on the development set of \scifact.}
\label{tab:abs_ret}
\vspace{-0.25cm}
\end{table}
Table~\ref{tab:abs_ret} reports R@3 and R@5 for abstract retrieval.
The oracle (first row) shows that most claims from the development set have fewer than 3 relevant abstracts and all have fewer than 5.
For comparison, we show the effectiveness of TF-IDF used by \citet{Wadden2020FactOF}.

We find that using BM25 results in a effectiveness improvement of around 10 points in comparison to the TF-IDF baseline.
Using T5 to rerank the top-20 abstracts retrieved from BM25 results in a 17-point improvement over baseline.

However, there is almost no difference in efficacy whether T5 was fine-tuned on \scifact or on MS MARCO MED.
This is potentially due to the relatively small size of the \scifact dataset and that MS MARCO MED data is not entirely relevant to the target task.
Hence, we use T5 trained on MS MARCO in the full pipeline experiments (Section~\ref{section:full_pipeline}).

\subsection{Sentence Selection}

\begin{table}[t]
\small
\centering{
\begin{tabular}{l|lll}
\toprule
Method & P & R & F1\\
\toprule
RoBERTa-large & 73.71 & 70.49 & 72.07 \\
\midrule
T5 & \textbf{79.29} & \textbf{73.22} & \textbf{76.14}\\
\bottomrule
\end{tabular}
}
\vspace{-0.1cm}
\caption{Comparison of different sentence selection methods on \scifact's development set.}
\label{tab:sent_ret}
\vspace{-0.25cm}
\end{table}
Table~\ref{tab:sent_ret} reports the precision, recall and F1 scores for the sentence selection task.

We find that T5 (MS MARCO) fine-tuned on \scifact outperforms the RoBERTa-large baseline fine-tuned on \scifact used by \citet{Wadden2020FactOF}.
This result together with those from Table~\ref{tab:abs_ret} demonstrates the effectiveness of the T5 model at selecting evidence in various levels of granularity.

\subsection{Label Prediction}

\begin{table}[t]
\small
\centering\centering\resizebox{0.49\textwidth}{!}{
\begin{tabular}{l|l|lll}
\toprule
Method & Label & P & R & F1\\
\toprule
 & \supports & 92.56 & 81.16 & 86.49 \\
RoBERTa-large & \noinfo & 74.82 & \textbf{92.86} & 82.87 \\
 & \refutes & 77.05 & 66.20 & 71.21 \\
\midrule
 & \supports & \textbf{93.13} & \textbf{88.41} & \textbf{90.71}\\
T5 & \noinfo & \textbf{85.25} & \textbf{92.86} & \textbf{88.89}\\
 & \refutes & \textbf{86.76} & \textbf{83.10} & \textbf{84.89}\\
\bottomrule
\end{tabular}
}
\vspace{-0.1cm}
\caption{Comparison of different label prediction models, on \scifact's development set.}
\label{tab:label_pred}
\end{table}

In Table~\ref{tab:label_pred}, we present label-wise precision, recall, and F1 scores for the label prediction task.
In the case of \supports and \refutes labels, the input to the model are gold rationales from cited abstracts.
The exception is for \noinfo labels, whose cited abstracts are available but no gold rationales exist.
In this case, we pick the two most similar sentences according to TF-IDF from each of these abstracts.

The results across all labels demonstrate that T5 fine-tuned on \scifact's label prediction task shows significant improvements over the baseline RoBERTa-large that was fine-tuned on FEVER followed by fine-tuning on \scifact's label prediction task. 
We believe some of this can be credited to T5's pretraining on a mixture of multiple tasks. 
Although this mixture does not include FEVER, it contains various other NLI datasets including MNLI \cite{williams-etal-2018-broad} and QNLI \cite{rajpurkar2016squad}.

\subsection{Full Pipeline}
\label{section:full_pipeline}

\begin{table}[t]
\small
\centering\centering{}{
\begin{tabular}{l|lll}
\toprule
 \multicolumn{4}{c}{Label Only} \\
 \midrule
Method & P & R & F1 \\
\toprule
(1) Oracle (\vs) & 90.97 & 67.46 & 77.47 \\
(2) Oracle (ours) & \textbf{92.70} & \textbf{78.95} & \textbf{85.27} \\
\midrule
(3) \vs & 55.31 & 47.37 & 51.03 \\
\midrule
(4) \vt (BM25) & \textbf{70.88} & 61.72 & \textbf{65.98}\\
(5) \vt (T5) & 65.07 & \textbf{65.07} & 65.07 \\
\bottomrule
\toprule
\multicolumn{4}{c}{Label+Rationale}\\
\midrule
Method & P & R & F1 \\
\toprule
(6) Oracle (\vs)  & 85.16 & 63.16 & 72.53\\
(7) Oracle (ours) & \textbf{88.76} & \textbf{75.60} & \textbf{81.65}\\
\midrule
(8) \vs & 52.51 & 44.98 & 48.45 \\
\midrule
(9) \vt (BM25) & \textbf{67.03} & 58.37 & \textbf{62.40}\\
\hspace{-4.5pt}(10) \vt (T5) & 61.72 & \textbf{61.72} & 61.72\\
\bottomrule
\end{tabular}
}
\vspace{-0.1cm}
\caption{Full pipeline abstract-level effectiveness on \scifact's development set.}
\label{tab:fp_al}
\end{table}

\begin{table}[t]
\small
\centering\centering{}{
\begin{tabular}{l|lll}
\toprule
\multicolumn{4}{c}{Selection Only}\\
\midrule
Method & P & R & F1 \\
\toprule
(1) Oracle (\vs) & 79.41 & 59.02 & 67.71  \\
(2) Oracle (ours) & \textbf{83.54} & \textbf{72.13} & \textbf{77.42}\\
\midrule
(3) \vs  & 52.46 & 43.72 & 47.69  \\
\midrule
(4) \vt (BM25) & \textbf{67.70} & 53.83 & 59.97 \\
(5) \vt  (T5) & 64.81 & \textbf{57.37} & \textbf{60.87}\\
\bottomrule
\toprule
\multicolumn{4}{c}{Selection+Label}\\
\midrule
Method & P & R & F1 \\
\toprule
(6) Oracle (\vs) & 71.32 & 53.01 & 60.82 \\
(7) Oracle (ours) & \textbf{78.16} & \textbf{67.49} & \textbf{72.43}\\
\midrule
(8) \vs   & 46.89 & 39.07 & 42.62 \\
\midrule
(9) \vt (BM25) & \textbf{63.92} & 50.82 & 56.62\\
\hspace{-4.5pt}(10) \vt  (T5)  & 60.80 & \textbf{53.83} & \textbf{57.10}\\
\bottomrule
\end{tabular}
}
\vspace{-0.1cm}
\caption{Full pipeline sentence-level effectiveness on \scifact's development set.}
\label{tab:fp_sl}
\end{table}

In Table~\ref{tab:fp_al}~and~\ref{tab:fp_sl}, we report the precision, recall and F1 scores of abstract-level evaluation and sentence-level evaluation respectively for full pipeline systems.

Rows 1, 2, 6, 7 present the scores in the oracle abstract retrieval setting, where gold evidence abstracts are provided to systems.
We see that our pipeline outperforms \vs by around 10 F1 points at both the abstract and sentence level. 
The improvements are even more significant in the Abstract\textsubscript{Label+Rationale} and Sentence\textsubscript{Selection+Label} evaluation settings (rows 6, 7 in Tables~\ref{tab:fp_al}~and~\ref{tab:fp_sl}, respectively) which require more from systems in terms of sentence selection and label prediction.

In rows 3-5 and 8-10, we report scores in the full pipeline setting where systems are also required to retrieve relevant abstracts.
We evaluate two full pipeline systems, one that used BM25 alone and another that used BM25 followed by T5 (MARCO) for abstract retrieval.
Both these systems outperform the baseline system \vs by about 14 F1 points.
This comes as no surprise seeing that our models displayed significant improvements along each of the three steps.

Notice that in Table~\ref{tab:abs_ret}, using T5 (MARCO) brings large gains in terms of R@3 of the BM25 baseline.
Yet in the case of the full-pipeline with these two abstract retrieval methods, we only notice comparable efficacy on the development set. 
We believe this might be linked to the relatively small size of the development set and choose to probe the \scifact hidden test set with both these systems.

\begin{table}[t]
\centering\centering{}{
\small
\begin{tabular}{l|lll}
\toprule
\multicolumn{4}{c}{Label Only} \\
\midrule
Method & P & R & F1\\
\toprule
(1) \vs & 47.5 & 47.3 & 47.4\\
(2) \vt (BM25) & 63.1 & 60.8 & 61.9 \\
(3) \vt (T5) & \textbf{63.6} & \textbf{66.2} & \textbf{64.9} \\
\bottomrule
\toprule
 \multicolumn{4}{c}{Label+Rationale}\\
\midrule
Method & P & R & F1\\
\toprule
(4) \vs & 46.6 & 46.4 & 46.5 \\
(5) \vt (BM25) & 60.3 & 58.1 & 59.2\\
(6) \vt (T5) & \textbf{61.5} & \textbf{64.0} & \textbf{62.7}\\
\bottomrule
\toprule
\multicolumn{4}{c}{Selection Only} \\
\midrule
Method & P & R & F1\\
\toprule
(7) \vs & 45.0 & 47.3 & 46.1  \\
(8) \vt (BM25) & 64.9 & 58.9 & 61.8 \\
(9) \vt (T5) & \textbf{66.2} & \textbf{63.5} & \textbf{64.8} \\
\bottomrule
\toprule
\multicolumn{4}{c}{Selection+Label}\\
\midrule
Method & P & R & F1\\
\toprule
\hspace{-4.5pt}(10) \vs & 38.6 & 40.5 & 39.5 \\
\hspace{-4.5pt}(11) \vt (BM25) & 58.3 & 53.0 & 55.5\\
\hspace{-4.5pt}(12) \vt (T5) & \textbf{60.0} & \textbf{57.6} & \textbf{58.8}\\
\bottomrule
\end{tabular}
}
\vspace{-0.1cm}
\caption{Full pipeline effectiveness on \scifact's test set.}
\label{tab:fp_test}
\vspace{-0.25cm}
\end{table}

From Table~\ref{tab:fp_test}, it is clear that in the hidden test set, both our systems outperform the baseline \vs, with evaluation aspects like Sentence+Label (rows 10-12) showing relative improvements of around 50\%. 
Comparing with respective scores in Table~\ref{tab:fp_al},~\ref{tab:fp_sl}, we also see no indication of overfitting.
We also note that abstract retrieval using the two-stage approach brings significant gains here  (rows 5, 11 vs 6, 12) unlike in the development set.
This shows that neural reranking, even though used in a zero-shot formulation, is critical to getting higher quality abstracts from the corpus $\mathcal{C}$ thereby improving effectiveness in later stages too.

\subsection{Verification of COVID-19 claims}
Finally we evaluate our most effective pipeline, \vt (T5), on the two sets of COVID-related claims.
We do this in a zero-shot paradigm in that we do not fine-tune our model on either of these sets.

In the COVID-19 \scifact set, for each claim $q$, we use \vt (T5) to predict evidence abstracts, $\hat{\mathcal{E}}(q)$. 
A $(q, \hat{\mathcal{E}}(q))$ pair is considered \textit{plausible} if at least half of the evidence abstracts in $\hat{\mathcal{E}}(q)$ are found to have reasonable rationales and labels. 
For 30/36 claims, we find that \vt (T5) provides plausible evidence abstracts.
These claims have reasonable labels and evidence rationales selected successfully from evidence abstracts.
This is in comparison to the 23/36 claims that \vs provides plausible evidence, demonstrating the effectiveness of our system in the zero-shot case.

\begin{table}[t]
\centering\centering\resizebox{0.49\textwidth}{!}{
\small
\begin{tabular}{l|lll}
\toprule
Method  & Accuracy & F1-Macro & F1-Binary\\
\toprule
LiarMisinfo & 61.5 & 59.2 & 82.8 \\
LM Debunker  & 75.4 & 69.8 & 83.1 \\
\midrule
\vt (T5)  & \textbf{78.2} & \textbf{73.2} & \textbf{83.8}\\
\bottomrule
\end{tabular}
}
\vspace{-0.1cm}
\caption{Label prediction effectiveness on COVID-19 Scientific claims}
\label{tab:leemisinfo}

\vspace{-0.25cm}
\end{table}

In the COVID-19 Scientific set, we compare the effectiveness of \vt with that of two baselines considered by \citet{lee2020misinformation}.
Table~\ref{tab:leemisinfo} reports the accuracy, the F1-Macro and the F1-Binary scores on the test set.
The F1-Binary score corresponds to the F1 score of the \refutes label, since debunking misinformation is critical.
Note that the LM Debunker baseline uses the average scores across a four-fold cross-validation on the test set, unlike \vt and LiarMisinfo.
We observe that \vt outperforms both the baselines in a zero-shot setting, without any in-task tuning like the LM Debunker.
The adaptability of \vt to both these new tasks with no additional training makes a strong case for the strength of our pipeline.

\section{Conclusion}
We introduced \vt, a novel pipeline for scientific claim verification that exploits a generation-based approach to abstract ranking, sentence selection and claim verification.
Such systems are of significance in this age of misinformation amplified by the COVID-19 pandemic.
We find that our system outperforms the state of art in the end-to-end task.
We note improvements in each of the three steps, demonstrating the importance of this generative approach as well as zero-shot and few-shot transfer capabilities.
Finally, we find that \vt generalizes to two new COVID-19 related sets with no tuning of parameters while maintaining high efficacy.

Yet, there is still a large gap between our system and an oracle.
Ideally, a system that performs scientific claim verification should possess additional attributes like:
\begin{itemize}[leftmargin=*]
    \item Numerical reasoning --- the ability to interpret statistical and numerical findings and ranges.
    \item Biomedical background --- the ability to leverage knowledge about domain-specific lexical relationships.
\end{itemize}
Future work that incorporate such attributes might be critical towards building higher quality scientific fact verification systems.

\section*{Acknowledgements}

This research was supported in part by the Canada First Research Excellence Fund and the Natural Sciences and Engineering Research Council (NSERC) of Canada.
Additionally, we would like to thank Google for computational resources in the form of Google Cloud credits.

\bibliography{main}
\bibliographystyle{acl_natbib}
\end{document}